\documentclass[conference]{IEEEtran}
\IEEEoverridecommandlockouts

\usepackage[utf8]{inputenc}
\usepackage[T1]{fontenc}
\usepackage{cite}
\usepackage{amsmath,amssymb,amsfonts}
\usepackage{graphicx}
\usepackage{textcomp}
\usepackage{xcolor}

\usepackage{mathtools}

\usepackage{pgf}
\usepackage{tikz}
\usetikzlibrary{arrows,automata}

\usepackage{algorithm}
\usepackage{algpseudocode}

\usepackage{csquotes}

\def\BibTeX{{\rm B\kern-.05em{\sc i\kern-.025em b}\kern-.08em
    T\kern-.1667em\lower.7ex\hbox{E}\kern-.125emX}}
\begin{document}

\title{Preventing the Generation of Inconsistent Sets of Classification Rules\\
}

\author{\IEEEauthorblockN{Thiago Zafalon Miranda}
\IEEEauthorblockA{\textit{Department of Computer Science} \\
\textit{Federal University of São Carlos}\\
São Carlos, Brazil \\
thiago@ufscar.br}
\and
\IEEEauthorblockN{Diorge Brognara Sardinha}
\IEEEauthorblockA{\textit{Department of Computer Science} \\
\textit{Federal University of São Carlos}\\
São Carlos, Brazil \\
diorge.sardinha@ufscar.br}
\and
\IEEEauthorblockN{Ricardo Cerri}
\IEEEauthorblockA{\textit{Department of Computer Science} \\
\textit{Federal University of São Carlos}\\
São Carlos, Brazil \\
cerri@ufscar.br}
}

\newcommand{\todo}[1]{\textcolor{red}{\texttt{#1}}}

\maketitle

\begin{abstract}
In recent years, the interest in 
interpretable classification models has grown.
One of the proposed ways to improve
the interpretability of a rule-based
classification model is to use sets
(unordered collections) of rules,
instead of lists (ordered collections) of rules.
One of the problems associated with sets
is that multiple rules may cover a single instance,
but predict different classes for it,
thus requiring a conflict resolution strategy.
In this work,
we propose two algorithms capable of 
finding feature-space regions inside which
any created rule would be consistent with
the already existing rules,
preventing inconsistencies from arising.
Our algorithms do not generate classification
models, but are instead meant to 
enhance algorithms that do so,
such as Learning Classifier Systems.
Both algorithms are described and analyzed
exclusively from a theoretical perspective,
since we have not modified a
model-generating algorithm to incorporate
our proposed solutions yet.
This work presents the novelty
of using conflict avoidance strategies
instead of conflict resolution strategies.
\end{abstract}


\section{Introduction}

Classification is one of the commonest
tasks of Machine Learning, 
concisely described in~\cite{otero2013improving}
as the generation of a model that
learns relations between predictive
features and target features.
This learning occurs by
adjusting the internal parameters
of the model.

In recent years,
the interest in interpretable classification models has grown,
partly due to regulations such as the General Data Protection Regulation (commonly known as GDPR),
that created a \enquote{right to explanation},
a regulation \enquote{whereby a user can ask for an explanation of an algorithmic decision that significantly affects them}~\cite{goodman2016european}.

Even though interpretability,
in the context of classification models,
is not an objectively and consistently defined concept~\cite{lipton2016mythos},
it is reasonable to say that
some types of classification models are inherently
more interpretable than others;
a Decision Tree~\cite{quinlan1986induction},
for instance,
can be said to be more interpretable than 
a Deep Neural Network~\cite{lecun2015deep}.

\begin{figure}
\centering
\includegraphics[width=0.8\columnwidth,keepaspectratio]{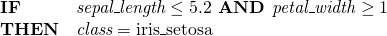}
\caption{Simple Classification Rule}
\label{fig:rule-example}
\end{figure}

It is generally accepted that 
rule-based classifiers are among 
the most interpretable~\cite{freitas2014comprehensible}.
The training phase of such classifiers 
usually consist in creating and
tuning a list of classification rules.
A classification rule usually has two components,
its antecedent and its consequent.
The antecedent is a collection of tests
over feature values,
and the consequent is the label\footnote{Or set of labels, in multi-label classification.}
that will be assigned to the
dataset instance which will be classified,
if it passes all the antecedent's tests. 
A simple classification rule is
exemplified in Figure~\ref{fig:rule-example}.

The interpretability of a rule-based classification model
is frequently measured by its size,
i.e. the number of rules in the model
and/or the number of features tested
by the rules~\cite{freitas2014comprehensible}.
Therefore, many algorithms that generate
interpretable classification models 
try to minimize the model size.

In~\cite{otero2013improving}, however,
the authors propose an alternative way
of improving the interpretability of
a rule-based classification model,
using sets (unordered collections) of rules, 
instead of lists (ordered collections).

If a classifier employs a list of rules,
then its n-th rule cannot be correctly interpreted alone,
because an instance that is covered by it
may also be covered by a previous rule;
the actual class predicted by the classifier
would be the one of the previous rule.
Using, instead, a set of rules allows the user
to analyze the rules individually,
making the model more interpretable.

However, using a set of rules may create conflicts
when multiple rules (with different consequents)
cover the same dataset instance.
The authors of~\cite{otero2013improving}
discuss two conflict resolution strategies,
allowing the classifier to function properly
even if it contains multiple rules that contradict each other.

In this work,
we propose two algorithms
capable of finding sets of
feature-space regions such
that any rule created within those regions
will always be consistent with $R$.
In this context,
a rule $r_1$ is said to be
consistent with a rule $r_2$ if
their consequents are identical or if
there is no intersection between their antecedents,
i.e. it is not possible to create an object
that would be covered by rules that predict
different labels.
A set of rules is said to be consistent
if each rule of the set is consistent 
with each other.

By only creating consistent rules,
one avoids the problem of conflicting
predictions entirely, hence
improving the interpretability
of the classification model.

The proposed algorithms do not generate
classification models by themselves,
instead they are meant to enhance
algorithms that do so.
They can be used, for instance,
during the initialization and mutation
phases of a genetic algorithm.
Since no model is directly generated,
our algorithms can only be 
evaluated by modifying an existing
model-building algorithm to use one
of the methods,
then measuring the relative change on the
induced models.
The two algorithms themselves
are independent of the metrics chosen.

It is interesting to note that
our algorithms are not
sensitive to the type of
the consequent of the rules,
as long as they can be tested for equality.
This means that they can be used
to supplement algorithms that
generate any kind of rule
format\footnote{
Such as hierarchical multi-label
or flat single-label rules.},
since inconsistencies
between rules arise from overlaps
between the rules' antecedents.

The remainder of this work is organized as follows:
in Section~\ref{sec:related_works} we discuss related works;
in Section~\ref{sec:proposed-algos} we present
two algorithms to aid the creation of consistent rules,
and analyze their properties;
in Section~\ref{sec:conclusion} we discuss future works
and present our conclusions.

\section{Related Works}
\label{sec:related_works}

The concept of interpretability has been
a point of contention in Artificial Intelligence (AI) literature.
There are many different views on what constitute
an interpretable classification model,
how to measure interpretability,
and whether it is necessary to, or even worth to,
sacrifice predictive power of a classifier in favor
of its interpretability.
Some authors have proposed mechanisms
to improve the interpretability of black-box models~\cite{caruana1999case},
while other have focused on transparent rule-based models,
such as Learning Classifier Systems~\cite{holland1999learning}.
In this section, we will discuss some of the works which 
have focused on algorithms that generate rule-based models,
and why interpretability is important.

In~\cite{varshney2018interpretability}
and~\cite{nirenburg2017cognitive}
the authors argue that AI models
do not usually operate in a vacuum,
they interact with humans,
and that various types
of Human-AI interactions may benefit
from an interpretable model.

In areas such as bioinformatics
(protein function prediction,
gene function prediction, among others)
it is important that the
classification model is interpretable,
in order to make it possible for its users 
to validate it~\cite{otero2010hierarchical}.
In medical and financial applications,
understanding a computer-induced model
is often a prerequisite for users to
trust the model's
predictions~\cite{freitas2014comprehensible}.

Considering that rule-based classification models are
inherently transparent, thus interpretable,
many algorithms that generate
interpretable models have been published
(see the discussion of transparency in~\cite{lipton2016mythos}).

The Decision Tree algorithm C4.5~\cite{quinlan2014c4},
for instance, generates models
that can be interpreted 
as easily as a flowchart.
It also employs a pruning strategy
that improves simultaneously the
interpretability of the model,
by reducing its size, and its
predictive power, by reducing
overfitting.

In~\cite{clare2001knowledge},
the authors modified the algorithm C4.5
to handle multi-label classification.
One of the most interesting parts
of their work, 
from an interpretability perspective,
was the generation of a set of rules
from the decision tree.
This process of \enquote{splitting}
a decision tree into a set of rules
is one of the few processes
that we know of that can generate a
consistent set of rules.

In~\cite{blockeel2002hierarchical},
the authors propose an algorithm 
based on Predictive Clustering Trees (PCTs)~\cite{blockeel1998top}
to perform hierarchical multi-label classification
using a single, global model.
PCT-based algorithms see decision trees as hierarchies
of clusters and as such, during the model training phase,
they try to minimize intra-cluster variance.
The proposed algorithm, called Clus-HMC,
was later modified in~\cite{vens2008decision}
to handle class hierarchies organized as
Directed Acyclic Graphs (DAGs),
and used in~\cite{schietgat2010predicting} to generate
a collection of trees which build an ensemble.

In~\cite{wang2005agent}, the
authors propose an evolutionary algorithm
to generate interpretable fuzzy classification rules
by using the Pittsburg approach,
in which each individual of the population
represents a complete classifier.
The fittest selection mechanism used was the multi-objective algorithm NSGA-2~\cite{deb2002fast},
and the functions being optimized were accuracy,
number of rules, and length of rules.
The authors also discuss the issue of
interpretability of fuzzy classification rules and strategies
to improve it, such as merging similar fuzzy sets.

In~\cite{cerri2012genetic},
the authors propose a Genetic Algorithm (GA)
to generate interpretable traditional (non-fuzzy)
classification rules.
The algorithm, called HMC-GA, is the only
GA-based method in the literature that
is capable of building a global hierarchical
multi-label classification model~\cite{gonccalves2018survey}.

In~\cite{parpinelli2002data},
the authors propose the first ant colony-based
classification algorithm, called Ant-Miner.
It generates lists of classification rules,
and had, in its original version,
the limitation of only handling categorical features. Ant-Miner was used as a base for many algorithms,
such as Multi-Label Ant-Miner (MuLAM)~\cite{chan2006new},
which generates flat (i.e. non-hierarchical) multi-label classification rules;
cAnt-Miner~\cite{otero2008cant},
which removed the restriction of using only categorical features;
h-Ant-Miner~\cite{otero2009hierarchical}, 
which generates hierarchical single-label classification rules;
and hm-Ant-Miner~\cite{otero2010hierarchical},
which generates hierarchical multi-label classification rules.

In~\cite{otero2013new}, 
the authors propose a new sequential covering strategy for cAnt-Miner,
in an algorithm called cAnt-Miner$_{\text{pb}}$.
This algorithm was later enhanced to generate sets (unordered collections) of rules
in an algorithm called Unordered cAnt-Miner$_{\text{pb}}$~\cite{otero2013improving}.
The authors of Unordered cAnt-Miner$_{\text{pb}}$ argue that
a set of rules 
is more interpretable
than a list (ordered collection) of rules.
They also propose a new 
interpretability metric, called
\textit{Prediction-Explanation Size},
that accounts for the inter-dependency
of rules in lists.

The sets of rules generated by
Unordered cAnt-Miner$_{\text{pb}}$
could contain inconsistent rules,
i.e. multiple rules that cover
the same dataset instance but predict
different classes for it.
The authors discuss mechanisms
to resolve such conflicts when they
arise (i.e. when making a prediction),
such as using the rule with the highest quality,
or aggregating the predictions
of the conflicting rules and selecting
the most common label in the aggregation.

The idea that sets of rules
are more interpretable than lists of rules,
and the fact that there are,
as far as we are aware,
no mechanisms to prevent the
generation of inconsistent rules,
motivated us to research and
develop the algorithms described
in Section~\ref{sec:proposed-algos}.

\section{Proposed Algorithms}
\label{sec:proposed-algos}

We propose two algorithms to solve the problem 
of adding a new rule to an existing set of consistent rules,
creating an expanded rule set that is still consistent.
More specifically,
the algorithms find feature-space
regions in which rules can be created
while being consistent with an already existing set of rules;
the actual creation of the rules inside such regions
is not within the scope of the methods.

There is a important distinction between identifying
if a new rule is consistent with an existing collection
of rules and creating a new rule that is consistent with an
existing collection of rules;
the first task,
analogous to determining if a cake tastes good,
is trivial;
the second task,
analogous to baking a good cake,
is far more complex.
The algorithms we propose are meant to guide
the execution of the later task.

To the best of our knowledge,
both algorithms,
and the conflict avoidance approach they employ,
are new to the literature.
We refer to these algorithms as 
Constrained Feature-Space Greedy Search (CFSGS)
and Constrained Feature-Space Box-Enlargement (CFSBE).
We present them, respectively,
in Section~\ref{subsec:CFSGS} and Section~\ref{subsec:BE}.

Whenever we refer to a rule,
unless stated otherwise,
we will be referring only to its antecedent.
The case in which rules have the same consequent
but different antecedents,
hence are consistent with each other, 
will be discussed in Section~\ref{sec:identical-consequents}.

We will assume that all predictive
features are continuous.
Both algorithms can handle categorical features,
but explaining them exclusively
in terms of continuous features
allows a better visualization and explanation.
We will discuss the treatment of categorical features
in Section~\ref{sec:categorical-features}.

In order to simplify the explanation of both
algorithms we will use the convention that rules
have exactly one \textit{feature test}
for each predictive feature
and have the format shown
in Equation~\ref{eq:rule_format},
where $test_i$ denotes the i-th test of the rule,
i.e. the test over the i-th feature,
and $|f|$ denotes the number of features
in the dataset.
We will also use the convention that feature tests have 
the format shown in Equation~\ref{eq:test_format},
where $f_i$ denotes the value of the i-th
feature of the dataset instance being tested,
and $test.lower$ and $test.upper$ are, respectively,
the lower and upper bound values of the test.

\begin{align}
\begin{split}
\label{eq:rule_format}
    rule := test_1 \wedge test_2 \wedge \dots \wedge test_{|f|}
\end{split}\\
\begin{split}
\label{eq:test_format}
    test := test.lower \leq f_i < test.upper
\end{split}
\end{align}

It is important to observe that 
the lower bound of a test is inclusive,
but the upper bound is exclusive.
This definition prevents inconsistencies
when two tests from different rules
\enquote{touch} each other,
i.e. the upper bound of the i-th test
from a rule has the same value 
as the lower bound of the i-th test
from another rule.
To exemplify this, consider the 
rules described in Equations~\ref{eq:consistent_rules1}
and~\ref{eq:consistent_rules2}.
If we use an inclusive upper bound,
a person with $age=10$ will be covered by both rules,
which will make the rules inconsistent with each other.

\begin{align}
rule_1 = \textbf{IF } & 0 \leq age < 10  & \textbf{ THEN } class=child 
\label{eq:consistent_rules1}\\
rule_2 = \textbf{IF } & 10 \leq age < 99 & \textbf {THEN } class=adult
\label{eq:consistent_rules2}
\end{align}

Both algorithms can be more easily explained if
we use a geometric interpretation,
that is, by viewing the antecedent
of a classification rule as
an $|f|$-dimensional hyperrectangle,
being $f$ the set of features of the dataset.

\begin{figure}
\centering
\includegraphics[width=0.7\columnwidth,keepaspectratio]{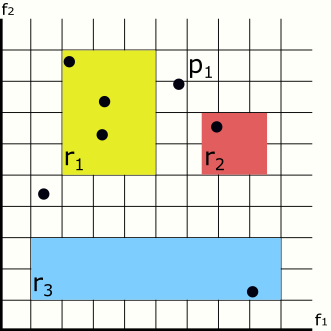}
\caption{Classification rules in 2D feature-space}
\label{fig:rules1}
\end{figure}

Figure~\ref{fig:rules1} shows what
three classification rules,
represented as colored rectangles,
and seven dataset instances,
represented as black dots,
could look like in a classification problem
with two predictive features, $f_1$ and $f_2$.

Considering that the grid squares in
Figure~\ref{fig:rules1} have unitary length, 
the antecedent of the rules depicted in the figure
can be formally described by
Equations~\ref{eq:rule1},
\ref{eq:rule2} and~\ref{eq:rule3}.

\begin{align}
\begin{split}
\label{eq:rule1}
rule_1 =
\begin{cases}
    test_{1} = 2 \leq f_1 < 5  \\
    test_{2} = 5 \leq f_2 < 9 
\end{cases}
\end{split}\\
\begin{split}
\label{eq:rule2}
rule_2 =
\begin{cases}
    test_{1} = 6.5 \leq f_1 < 8.5  \\
    test_{2} = 5 \leq f_2 < 7 
\end{cases}
\end{split}\\
\begin{split}
\label{eq:rule3}
rule_3 =
\begin{cases}
    test_{1} = 1 \leq f_1 < 9  \\
    test_{2} = 1 \leq f_2 < 3 
\end{cases}
\end{split}
\end{align}

The premise of the two proposed algorithms
is that if we can find a region of the
feature-space that is not covered
by any rule, we could create
a rule inside such region.
The new rule would be consistent
with the already existing rules,
because no dataset instance could be
simultaneously covered by more than one rule.

\subsection{Constrained Feature-Space Greedy Search}
\label{subsec:CFSGS}

Since the region covered by a rule
is the region described by the
conjunction of its tests,
the region \textit{not covered} by it
can be described by the
the disjunction of the negation
of its tests, as we can see 
in Equations~\ref{eq:rule-negation}
and~\ref{eq:test-negation}.

\begin{align}
\label{eq:rule-negation}
rule       &= test_1 \wedge \dots \wedge test_n \\ \nonumber 
\lnot rule &= \lnot (test_1 \wedge \dots \wedge test_n) \\ \nonumber
\lnot rule &= (\lnot test_1) \lor \dots \lor (\lnot test_n) \\ \nonumber
\end{align}
\begin{align}
\label{eq:test-negation}
test_i &= lower_i \leq f_i < upper_i\\ \nonumber 
test_i &= (lower_i \leq f_i) \land (f_i < upper_i)\\ \nonumber
\lnot test_i &= \lnot((lower_i \leq f_i) \land (f_i < upper_i)) \\ \nonumber
\lnot test_i &= \lnot(lower_i \leq f_i) \lor \lnot(f_i < upper_i) \\ \nonumber
\lnot test_i &=  (lower_i > f_i) \lor (f_i \geq upper_i) \\ \nonumber
\lnot test_i &=  (f_i < lower_i) \lor (f_i \geq upper_i) \nonumber
\end{align}

Negating the tests of $r_1$, for instance,
results in four inequalities (or constraints),
each describing a region not covered by the rule:

\begin{itemize}
    \item $f_1 < 2$  , the region to the left of the yellow rectangle
    \item $f_1 \geq 5$ , the region to the right of the yellow rectangle
    \item $f_2 < 5$ , the region below the yellow rectangle
    \item $f_2 \geq 9$ , the region above the yellow rectangle
\end{itemize}

In Constrained Feature-Space Greedy Search (CFSGS),
we create a collection $C$ with
the constraints generated by the negation
of all the tests from all the rules,
with $C_{i,j}$ denoting the j-th
inequality generated from the i-th rule.
The constraints generated from the
negation of the tests of $r_1$, $r_2$, and $r_3$
are shown in Table~\ref{tab:constraints_table}.
As we assumed that all features are continuous,
the number of constraints generated per rule,
$n_f$, is equal to $2 \cdot |f|$.
We discuss the relation between
$n_f$ and the data type of the features
(categorical or continuous)
in Section~\ref{sec:categorical-features}.

\begin{table}[htbp]
\caption{Constraints Generated From Rules}
\small
\centering
\caption{Constraints Generated From Rules}
\begin{tabular}{c|c}
    Index in $C$ & Constraint Generated \\  \hline
    $C_{1,1}$ & $f_1 < 2$ \\ 
    $C_{1,2}$ & $f_1 \geq 5$ \\
    $C_{1,3}$ & $f_2 < 5$ \\
    $C_{1,4}$ & $f_2 \geq 9$ \\
    
    $C_{2,1}$ & $f_1 < 6.5$ \\
    $C_{2,2}$ & $f_1 \geq 8.5$ \\
    $C_{2,3}$ & $f_2 < 5$ \\
    $C_{2,4}$ & $f_2 \geq 7$\\
    
    $C_{3,1}$ & $f_1 < 1$ \\
    $C_{3,2}$ & $f_1 \geq 9$ \\
    $C_{3,3}$ & $f_2 < 1 $ \\
    $C_{3,4}$ & $f_2 \geq 3$ \\
\end{tabular}
\label{tab:constraints_table}
\end{table}

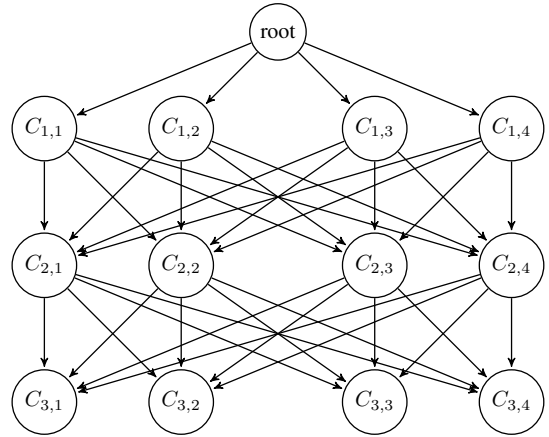
\begin{figure}

\centering

\vspace*{2.5mm}
\resizebox {0.8\columnwidth}{!}{%

\begin{tikzpicture}[->,>=stealth',shorten >=1pt,auto,node distance=2.2cm,
                    semithick]
  \tikzstyle{every state}=[fill=white,draw=black,text=black]

  \node[state] (ROOT)                    {root};
  
  \node[state] (R12) [below left of=ROOT]   {$C_{1,2}$};
  \node[state] (R13) [below right of=ROOT]  {$C_{1,3}$};
  \node[state] (R11) [left of=R12]          {$C_{1,1}$};
  \node[state] (R14) [right of=R13]         {$C_{1,4}$};
  
  \node[state] (R21) [below of=R11]  {$C_{2,1}$};
  \node[state] (R22) [below of=R12]  {$C_{2,2}$};
  \node[state] (R23) [below of=R13]  {$C_{2,3}$};
  \node[state] (R24) [below of=R14]  {$C_{2,4}$};
  
  \node[state] (R31) [below of=R21]  {$C_{3,1}$};
  \node[state] (R32) [below of=R22]  {$C_{3,2}$};
  \node[state] (R33) [below of=R23]  {$C_{3,3}$};
  \node[state] (R34) [below of=R24]  {$C_{3,4}$};
  
  \path (ROOT) edge node {} (R11)
               edge node {} (R12)
               edge node {} (R13)
               edge node {} (R14)
               
        (R11) edge node {} (R21)
              edge node {} (R22)
              edge node {} (R23)
              edge node {} (R24)
              
        (R12) edge node {} (R21)
              edge node {} (R22)
              edge node {} (R23)
              edge node {} (R24)
              
        (R13) edge node {} (R21)
              edge node {} (R22)
              edge node {} (R23)
              edge node {} (R24)
              
        (R14) edge node {} (R21)
              edge node {} (R22)
              edge node {} (R23)
              edge node {} (R24)
              
        (R21) edge node {} (R31)
              edge node {} (R32)
              edge node {} (R33)
              edge node {} (R34)
              
        (R22) edge node {} (R31)
              edge node {} (R32)
              edge node {} (R33)
              edge node {} (R34)
              
        (R23) edge node {} (R31)
              edge node {} (R32)
              edge node {} (R33)
              edge node {} (R34)
              
        (R24) edge node {} (R31)
              edge node {} (R32)
              edge node {} (R33)
              edge node {} (R34);
\end{tikzpicture}%
}
\caption{Constraints as DAG}
\label{fig:dag}
\end{figure}

By using Algorithm~\ref{alg:dagbuild} 
we can organize $C$ into a Directed Acyclic Graph (DAG).
Doing so allows us to perform a
greedy search to find all subsets of $C$
that contain exactly one constraint from each rule
and all constraints are simultaneously satisfiable.
Such subsets of $C$ describe non-empty regions
of the feature-space that are not covered by any rule.
The DAG generated from the constraints
of $r_1$, $r_2$, and $r_3$ is shown in
Figure~\ref{fig:dag}.

\begin{algorithm}
\caption{CFSGS - DAG Build}
\begin{algorithmic}[1]
\small

\Require
\Statex Collection of constraints: $C$
\Statex Number of existing rules: $n_r$
\Statex Number of constraints generated from a rule: $n_f$ (number of categorical features plus 2 times the number of continuous features)

\Ensure
\Statex The DAG $G$ representing the constrained feature-space to be explored

\item[]

\State G $\gets$ ~New Directed Graph
\State $G.E \gets \{ \} $
\Comment The set of edges of G
\State $G.V \gets \{ root \}$
\Comment The set of vertices of G

\For {$i \in \{1, 2, \ldots , n_r\}$}
\For {$j \in \{1, 2, \ldots , n_f\}$}
    \State $G.V \gets G.V \cup \{{C_{i,j}}$\}
    \If {$i > 1 $}
    \For {$k \in \{1, 2, \ldots , n_f\}$}
        \State $G.E \gets G.E \cup \{ (C_{i-1,k} , C_{i,j}) \} $
    \EndFor
    \EndIf
\EndFor
\EndFor

\State \Return $G$

\end{algorithmic}

\label{alg:dagbuild}
\end{algorithm}

The objective of the search 
is to find \textit{consistent paths}
from the root to a leaf node.
A path is said to be consistent
\textit{iff} the constraints
represented by its nodes
can all be simultaneously
satisfied, such as the one described in 
Equation~\ref{eq:validpath}.
If adding a node to the
path currently being explored
makes the constraints unsatisfiable,
the search algorithm backtracks and tries adding another node.

\begin{align}
\{ C_{1,1}, C_{2,1}, C_{3,1} \} 
&= (f_1 < 2) \wedge (f_1 < 6.5) \wedge (f_1 < 1) 
\label{eq:validpath} \\
&= f_1 < 1 \nonumber
\end{align}

We could fully explore the DAG to generate
all consistent paths, but since we only need one,
we stop the search as soon as the first one is found.
It is important to observe that the order
in which the nodes within a level are explored
determine the order in which paths are found.
If the nodes are explored in a
lexicographic order, e.g. $C_{1,1}$ 
is explored before $C_{1,2}$,
the first paths found will have a bias for the lower 
regions of the first features
(e.g. bottom left area in Figure~\ref{fig:rules1}).
To remove this bias, it suffices to explore
the nodes of each level in a random order.

The runtime complexity of building the DAG
can be expressed in function of the number
of existing rules $n_r$ and the number of
constraints generated per rule $n_f$.

In Algorithm~\ref{alg:dagbuild},
line 7 is executed $n_r \cdot n_f$ times,
due to the two nested loops.
Line 10 is executed $n_r \cdot (n_f - 1) \cdot n_f$ times,
since the conditional in line 8 decreases
the number of iterations by one.
Considering the cost of line 7 to be a constant $k_1$,
and the cost of the line 10 to be a constant $k_2$,
the total cost of building the DAG is described by
Equation~\ref{eq:cost-cfsgs}.

\begin{align}
\label{eq:cost-cfsgs}
    T(n_f, n_r) &= n_r \cdot n_f \cdot k_1 + n_r \cdot (n_f - 1) \cdot n_f \cdot k_2 \\
    T(n_f, n_r) &\in \mathcal{O}(n_f^\text{\hspace{1.5mm}2}\text{\hspace{1mm}} \cdot n_r)
    \nonumber
\end{align}

For the search algorithm,
the worst-case scenario of having
no possible consistent paths
would cause the exploration of every path.
Since creating a path is choosing an edge,
out of $n_f$ edges,
repeated over $n_r$ levels of the graph,
there are exactly $n_f^{\text{\hspace{1.5mm}}n_r}$ possible paths,
leading to a complexity cost of
$\mathcal{O}(n_f^{\text{\hspace{1.5mm}}n_r})$.

The total computational cost of CFSGS
is the sum of building the DAG and exploring it,
which makes the method bounded by
$(\mathcal{O}(n_f^\text{\hspace{1.5mm}2}\text{\hspace{1mm}} \cdot n_r)
+ \mathcal{O}(n_f^{\text{\hspace{1.5mm}}n_r}))
\in \mathcal{O}(n_f^{\text{\hspace{1.5mm}}n_r})$.
While the algorithm is capable of finding all sub-regions 
where a consistent rule could be created,
this exponential complexity cost
makes it unsuitable for many applications.

\subsection{Constrained Feature-Space Box-Enlargement}
\label{subsec:BE}

Extending the geometrical interpretation
provided by CFSGS,
we would like a way
to guide the search through the DAG such that
the cost becomes polynomial in relation to
the number of features and rules.
Constrained Feature-Space Box-Enlargement's 
central idea
is that it is possible to leverage
information from the training dataset
to visit only nodes that
lead to a possible consistent path to a leaf node.

Instead of searching the feature-space
for suitable regions,
CFSBE starts from a point known not to be
covered by any rule,
called a \enquote{seed},
and~\enquote{enlarge} this point along
the different dimensions,
creating a box that does not
overlap with any of the existing
rules' antecedents.
A rule created inside such box
would also not overlap with
the existing rules,
so they would be consistent,
regardless of their consequent.

Searching for arbitrary non-covered points
is equivalent to searching for
non-covered regions,
being as computationally expensive as CFSGS.
However, if we keep track of which dataset
instances are not covered by any rules
we may use one of such instances
as the seed.
By using an associative array structure
to map which points are covered
by which rules,
choosing a seed would have cost
$\mathcal{O}(1)$,
and updating the structure
on the insertion or removal of a rule
would have cost
$\mathcal{O}(n)$,
$n$ being the number of instances
in the training dataset.

\begin{figure}
\centering
\includegraphics[width=0.7\columnwidth,keepaspectratio]{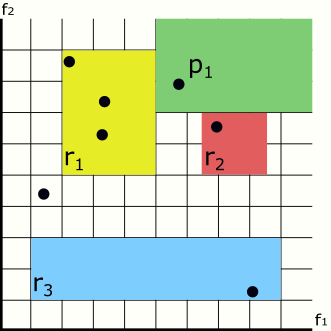}
\caption{CFSBE - Horizontal Axis First}
\label{fig:rules2}
\end{figure}

\begin{figure}
\centering
\includegraphics[width=0.7\columnwidth,keepaspectratio]{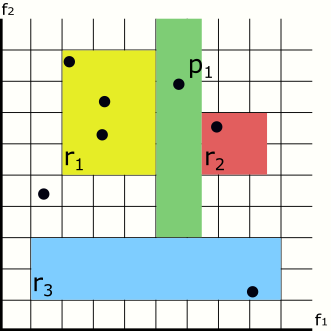}
\caption{CFSBE - Horizontal Axis First}
\label{fig:rules3}
\end{figure}

After selecting a non-covered point as seed,
we must choose the order in which
the dimensions will be expanded.
Consider the point $p_1$
in Figure~\ref{fig:rules1};
if we first enlarge it along the
$f_1$ dimension and $f_2$ afterwards,
we end up with the green region
shown in Figure~\ref{fig:rules2}.
If, however, we start with $f_2$,
we end up with the green region
shown in Figure~\ref{fig:rules3}.



Once a seed point and a feature order is chosen,
a degenerate hyperrectangle
(or box) is created around the point.
The box is then enlarged in each
dimension according to the chosen ordering.

To determine the limit of the expansion
along each dimension, that is,
the boundaries to which the box can grow
without overlapping with existing rules,
it is necessary to check which rules
\enquote{intersect} with the box 
on the other dimensions.

To exemplify the need for
checking for \enquote{intersections} on
dimensions that are not currently
being expanded on, 
consider the point $p_1$ in
Figure~\ref{fig:rules1}.
Even though it is not covered by
any rule, it does pass
the $f_2$ feature test of 
the rule $r_1$, i.e. it \enquote{intersects}
the rule $r_1$ on the $f_2$ dimension.
If we create our box around $p_1$ and grow it 
along the $f_1$ dimension,
it would eventually contain points that satisfy both
tests of $r_1$, i.e. there would be an intersection
between the box and $r_1$.

The method used to safely grow a box in such a
way that it does not overlap with the boxes
described by the antecedents of a set of rules
is presented in Algorithm~\ref{alg:boxenlarge}.

\begin{algorithm}
\caption{Constrained Feature-Space Box-Enlargement}
\begin{algorithmic}[1]
\small

\Require
\Statex Set of non-overlapping rules $R$
\Statex Seed point $x$, not covered by any rule of $R$
\Statex Order of dimensions to expand $O_d$ (a permutation of $\{1, 2, \ldots, |O_d|\}$)

\Ensure
\Statex The box $B$ covers a possible hyperrectangle containing $x$ that can not be further expanded, and does not intersect any rule of R

\item[]

\State $B \gets$ ~Degenerate rectangle with $|O_d|$ dimensions

\For{$i \in O_d$}
    \State $B_i.lower \gets x_i$
    \State $B_i.upper \gets x_i$
\EndFor

\For{$d \in O_d$}
    \State $B_d.lower \gets -\infty$
    \State $B_d.upper \gets +\infty$
    
    \For {$r \in R$}
        \If {$Intersects(B, r, d)$}
            \If {$x_d \geq r_d.upper$}
                \State $B_d.lower \gets max(B_d.lower,~r_d.upper)$
            \Else
            \Comment{Meaning $x_d < r_d.lower$, otherwise $x$ would be covered by $r$}
                \State $B_d.upper \gets min(B_d.upper,~r_d.lower)$
            \EndIf
        \EndIf
    \EndFor
    
\EndFor

\State\Return $B$

\item[]

\Function {Intersects}{Box B, Rule r, Dimension to skip d}

    \State $NumDimensions \gets \lvert r \rvert$
    \For{$i \in \{1, 2, \ldots, NumDimensions\} \setminus d$ }
        \State $u \gets B_i.upper > r_i.lower$
        \State $l \gets B_i.lower < r_i.upper$
        \If{$\lnot (u \land l)$}
            \State\Return False
        \EndIf
    \EndFor
    \State\Return True
\EndFunction

\end{algorithmic}
\label{alg:boxenlarge}
\end{algorithm}

It is worth noting that even though CFSBE
cannot find all suitable hyperrectangles,
it can find all the arguably relevant ones.
Consider the dataset
depicted in Figure~\ref{fig:rules1},
CFSGS could find a region below
the blue rectangle, while CFSBE cannot;
but since there are no instances there,
it is arguable that the rules created in
such region would not be useful,
as their coverage would be zero
and their predictive power could
not be measured on the dataset
used to train the classification model.

CFSBE's runtime complexity
can be calculated in a straightforward manner.
Let $d$ be the number of dimensions (features)
in the dataset and $r$ be the number
of rules in the set of existing rules.
The innermost part of the algorithm,
in lines 11 through 15,
has a constant cost, $k_1$,
because they do not depend on the values
of $d$ or $r$.
The $Intersects$ function has a loop
that executes at most $d - 1$ steps.
The \textit{contents} of this loop do not
depend on $d$ nor $r$,
therefore they also 
have a constant cost, $k_2$,
hence the worst-case scenario for this function
is $T_1(d) = (d-1) \cdot k_2$.

Lines 1 through 4 perform the
initialization of the degenerated rectangle,
which occurs in a loop with $d$ steps,
each step having constant cost $k_3$. 
The rest of the algorithm is a trivial
nesting of loops,
the first of which takes $d$ steps,
the second $r$ steps,
and the third,
inside the $Intersects$ function,
takes at most $d - 1$ steps,
as discussed previously.

The total cost of CFSBE
is described as $T_2$ in Equation~\ref{eq:cost-be}.
There are three terms in this summation,
the first being the creation of the degenerated rectangle,
the second the main algorithm body,
and the third value, $n$, comes from
keeping track of the available seeds,
as discussed previously.
Many algorithms that generate rule-based
classification models, however,
already have to create a mapping between
dataset instances and rules that cover them;
Learning Classifier Systems, for instance,
may need such information to calculate
the fitness of the rules~\cite{cerri2012genetic}.
Therefore, in practice,
the cost of CFSBE could be considered as
$\mathcal{O}(d^2 \cdot r)$.

\begin{align}
\label{eq:cost-be}
    T_2(d, r, n) &= d \cdot k_3 + d \cdot r\cdot T_1(d) \cdot k_1 + n \\
    T_2(d, r, n) &\in \mathcal{O}(d^2 \cdot r + n)
    \nonumber
\end{align}

\subsection{Tests Over Categorical Features}
\label{sec:categorical-features}

We explained both algorithms assuming
that the dataset contained
only continuous features,
but both algorithms can be modified
to handle categorical features.
Since a feature test over
a categorical feature is simply
an equality test,
the main difference for CFSGS
is that during the creation
of the collection of constraints $C$,
a single constraint is generated
for categorical features,
instead of two,
as we can see in
Equations~\ref{eq:test-cat1} and
\ref{eq:test-cat2}.
Consequently, the parameter
$n_f$ of Algorithm~\ref{alg:dagbuild}
equals to the number of categorical
features plus two times the number
of continuous features.

\begin{align}
\begin{split}
\label{eq:test-cat1}
rule_1 &=
\begin{cases}
    test_{1} = 2 \leq f_1 < 5  \\
    test_{2} = f_2 = yellow
\end{cases}
\end{split}\\
\begin{split}
\label{eq:test-cat2}
C &=
\begin{cases}
    C_{1,1} = f_1 < 2 \\
    C_{1,2} = f1 \geq 5\\
    C_{2,1} = f2 \neq yellow
\end{cases}
\end{split}
\end{align}

For CFSBE to handle categorical features,
we must change the data structure
that represents the enlarging box.
Instead of being a simple
associative array that maps
feature indices to continuous ranges,
it must now map feature indices to either
sets of values, for categorical features,
or continuous ranges,
for continuous features.
Considering this difference,
it is more appropriate to call
the Box a conflict-free \enquote{Region}.

Modifying the algorithm to check
whether the Region and a rule intersect
along a dimension that represents
a categorical feature is rather simple,
one only needs to check if the
value being tested by the rule
is a member of the set of values
of the Region for that dimension.
Similarly,
adjusting the Region's values
for a categorical dimension,
in order to avoid overlapping with a rule,
equates to removing the
value which is tested by the rule
from the set of values for that dimension.

\subsection{Rules With Identical Consequents}
\label{sec:identical-consequents}

We explained both algorithms using
the simplification of ignoring
the case in which the created rule
could overlap with rules that already
existed because they have the same consequent.
If the consequent of the rule that will be created
is known beforehand, then both CFSGS and CFSBE 
can be modified to allow rules with
the same consequent to overlap.

If, however, the consequent is not known
beforehand, to ensure that the
the rule created inside the region found
will be consistent with the already
existing rules, both CFSGS and CFSBE
must assume that the consequent will
be different from the consequents
of the rules that already exist,
i.e. that rules cannot overlap.
It is common for evolutionary algorithms
to generate the consequent of the rule
in function of its antecedent,
e.g.~\cite{cerri2012genetic, parpinelli2002data, chan2006new}.
In such cases, both our algorithms
will not allow intersections in the
antecedents.

Remember that rules are inconsistent,
and therefore require a conflict resolution strategy,
\textit{iff} their antecedents 
overlap but have different consequents,
i.e. they predict different labels for a
single dataset instance.
That means that if two rules have the same consequent,
then they are consistent, and don't require
a conflict resolution strategy, even if they overlap.

For CFSGS that means that during the creation
of $C$, the collection of constraints, it is
not necessary to generate constraints from rules
that have the same consequent as the rule that
will be created.
Not generating constraints from a rule allows
the region found by CFSGS to overlap with such
rule.

For CFSBE, when enlarging the box,
we can safely ignore rules that
have the same consequent as the rule that
will be created, again resulting in
the possibility of overlaps.
That can be achieved by either 
changing line 22 to skip such rules, 
or by simply removing them from the argument $R$.
\section{Conclusion and Future Work}
\label{sec:conclusion}

In this work we discussed the problem
of generating sets of rules without
inconsistencies and proposed 
two algorithms to solve this problem,
called CFSGS and CFSBE.

CFSGS is able to search through the
feature-space for any region
where the antecedent of a rule can be created
without creating inconsistencies with any existing rule.
However, the algorithm is computationally expensive.

The CFSBE algorithm, on the other hand,
can only find regions around dataset instances
that are not covered by any existing rule,
but its computational cost is far more reasonable.
We argue that the non-covered dataset instance 
requirement of CFSBE is not a hindering issue.

Neither algorithm is particularly useful by itself,
since both are meant to supplement
algorithms that generate rule-based classification models.
In the future, we intend to modify a Learning Classifier System
to use CFSBE both during the initial population creation and
during the mutation phases,
in order to study its effects on the predictive power
and interpretability of the generated models,
and whether it makes the models more prone to overfitting.

\section{Acknowledgments}
This study was financed in part by the Coordenação de Aperfeiçoamento de Pessoal de Nível Superior - Brasil (CAPES) - Finance Code 001.
R. Cerri thanks São Paulo Research Foundation (FAPESP) for the grant \#2016/50457-5.

\section{Conflicts of Interest}
The authors declare no conflict of interest. 
The financing instutions had no role in the design of the study,
in the writing of the manuscript 
or in the decision to publish the results.

\bibliographystyle{IEEEtran}
\bibliography{main}

\end{document}